\documentclass[5pt]{article}

\usepackage[letterpaper]{geometry}
\usepackage{spconf,amsmath,epsfig}
\usepackage{enumitem}

\usepackage{fancyhdr}
\begin{document}\sloppy

\def\x{{\mathbf x}}
\def\L{{\cal L}}

\title{Progressive refinement: a method of coarse-to-fine image parsing using stacked network}
%
%
%
%

\twoauthors
  {Jiagao Hu, Zhengxing Sun\thanks{This work is supported by National High Technology Research and Development Program of China (No. 2007AA01Z334); National Natural Science Foundation of China (No. 61321491, 61272219); Innovation Fund of State Key Laboratory for Novel Software Technology (No. ZZKT2013A12, ZZKT2016A11); Program for New Century Excellent Talents in University of China (No. NCET-04-04605); Nanjing University Innovation and Creative Program for PhD candidate (No. 2016013).}}
	{State Key Lab for Novel Software Technology,\\
    Nanjing University,
    Nanjing, China\\
    szx@nju.edu.cn}
  {Yunhan Sun, Jinlong Shi}
	{School of Computer Science and Engineering\\
	Jiangsu University of Science and Technology\\
	Zhenjiang, China}


\maketitle

\begin{abstract}
To parse images into fine-grained semantic parts, the complex fine-grained elements will put it in trouble when using off-the-shelf semantic segmentation networks. In this paper, for image parsing task, we propose to parse images from coarse to fine with progressively refined semantic classes. It is achieved by stacking the segmentation layers in a segmentation network several times. The former segmentation module parses images at a coarser-grained level, and the result will be feed to the following one to provide effective contextual clues for the finer-grained parsing. To recover the details of small structures, we add skip connections from shallow layers of the network to fine-grained parsing modules. As for the network training, we merge classes in groundtruth to get coarse-to-fine label maps, and train the stacked network with these hierarchical supervision end-to-end. Our coarse-to-fine stacked framework can be injected into many advanced neural networks to improve the parsing results. Extensive evaluations on several public datasets including face parsing and human parsing well demonstrate the superiority of our method.
\end{abstract}
\section{Introduction}
\label{sec:intro}

\begin{figure*}[t]
\centering
\includegraphics[width=125mm]{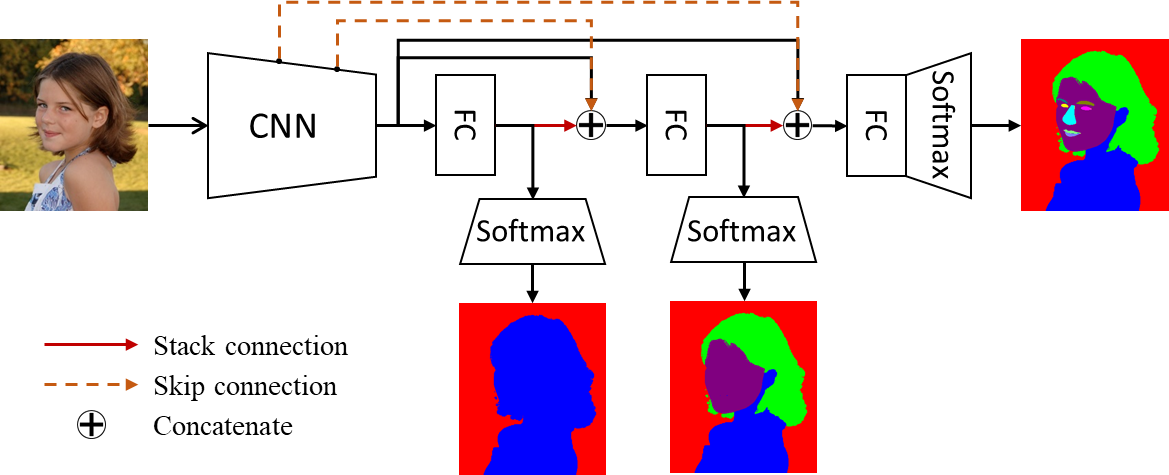}
\caption{Overview of the proposed method. The FC modules are the last layers in an FCN which output the prediction score maps, and the Softmax modules are used to restore the resolution of the score maps and get the pixel-wise labels using softmax. Best viewed in color}
\label{fig:framework}
\end{figure*}

Image parsing aims to segment an image into semantic regions and corresponding elements with more fine-grained semantics, which can provide full understanding of image contents. An effective image parsing facilitates various higher-level computer vision applications, such as image editing~\cite{Li2017b}, image retrieval~\cite{Liang2016b}, and artistic rendering~\cite{Yang2017b}.

Recently, fully convolutional networks (FCN)~\cite{Long2015a} based methods have achieved great success in various pixel-wise prediction tasks, such as semantic segmentation~\cite{Chen2017} and salient object detection~\cite{Wang2017b}. Image parsing can usually be cast as semantic segmentation problem with both tasks are to label every single pixel in the image. However, image parsing requires to segment images into more fine-grained semantic parts than general semantic segmentation tasks do. Due to the complex contextual of fine-grained parts, it will exhibit clear limitations when using the segmentation network to parse images. The accurate image parsing relies on the prior information of semantic part context and the details of semantic part structure. And the major issue for current FCN based models is lack of suitable strategy to utilize global contextual clues~\cite{Zhao2016a}. Thus, existing image parsing works often focus on the ways to complement the global context. For example, Li et al.~\cite{Li2017b} proposed to input addition shape prior of semantic parts as global constraint. Liang et al.~\cite{Liang2015} proposed to append LSTM layers to the network to capture long-distance global information from the whole image. To make good use of global image-level priors, Zhao et al.~\cite{Zhao2016a} proposed to use the pyramid pooling module to collet levels of information from multiple scales. In a word, most of existing methods either use prior knowledge of semantic parts as additional input, or aggregate the whole image-level clues to provide the global context for pixel-level image parsing.
Considering the image parsing process of human, people often firstly figure out the object in an image, and then gradually recognize the detail parts based on the context about that object. Inspired by this, an alternative strategy can be introduced to parse images progressively with refined semantic classes to incorporate part context. Before the final fine-grained parsing, an image is first parsed coarsely discarding the small elements to provide coarse-grained context. This can be achieved by stacking several segmentation networks with the former one parsing images coarsely and the latter one parsing images finely, while the former result is transmitted to the latter network as additional input, just like that in image synthesis~\cite{Zhang2016c} and salient object detection~\cite{Wang2017b}. However, stacking full networks crudely is a waste of computing resources and can make the network hard to be trained.

In this paper, a coarse-to-fine image parsing framework is proposed to parse images progressively with refined semantic classes by stacking several FCNs. The first network is trained to segment images at a coarse-grained level, and the last one is trained at the finest-grained level. To remove the redundant computation in stacked networks, we propose to share the image encoding parts (i.e., the former layers) of each network, which results in a standard FCNs with multiple stacked segmentation modules. To parse the fine-grained semantic parts precisely, we add skip connections from shallow layers which can provide more structural and localization details to the fine-grained segmentation modules. For the network training, we merge some classes in the finest-grained groundtruth label map to get a set of coarse-to-fine label maps, and train the network with this hierarchical supervision. The stacked network can be trained end-to-end and can get progressively refined hierarchical parsing result in a single forward pass. In addition, the coarse-to-fine stacking strategy can be injected into many advanced image segmentation networks for image parsing tasks. Fig.~\ref{fig:framework} shows the framework of our stacked networks.

The contributions of this paper are as follows:

\begin{itemize}
  \item We propose to parse images starting from a coarse-grained to the final fine-grained with gradual refinement by a generic stacked framework. With the contextual information from the coarse-grained result, the fine-grained parsing can be done better.
  \item We propose to utilize the details and localization information in the shallow layers to promote the fine-grained parsing by adding skip connections.
  \item We propose to generate hierarchical label maps by merging classes in groundtruth, and train the stacked network by hierarchical supervision based on the merged coarse-to-fine label maps.
\end{itemize}

\section{Methology}

\subsection{Stacked networks for image parsing}

When human observing an image, people usually firstly figure out the prominent object in it, and then recognize the detail parts based on prior knowledge about that object, which is a coarse-to-fine parsing process. Inspired by this process, we can use several FCNs to segment images progressively with refined semantic classes for image parsing task. The most intuitive way is to stack full FCNs one after another, with the output of the former FCN combined with the raw image feeding to the latter FCN as input, just as done in~\cite{Zhang2017c}. But in fact, the raw image is re-fed to same modules multiple times by doing this, which is a waste of computing resources. And that the coarse parsing result from former FCN must pass a full deep network, which could lead to information attenuating due to the long computation chain~\cite{Liang2015}.

In this paper, we propose to stack only the prediction layers on top of each other rather than the whole network, as shown in Fig.~\ref{fig:framework}. The FC module is the last convolutional layers in an FCN, which output a score map with \emph{C} channel (\emph{C} is the number of classes in the corresponding parsing granularity), just like the fully connected layers in a classification network. The corresponding pixel-wise segmentation can be obtained by upsampling the score map of each FC module to match the size of input and then feeding it to a soft-max classifier. In fact, this framework is similar to stacking several FCNs with shared former layers (i.e., layers before the FC modules of each FCN are shared). By doing so, we only need to feed the raw image to the network one time. Also, since the score map at the coarser-grained level is directly fed to the FC module at the finer-grained level, information from the coarse-grained level can be retained better.

\begin{figure}
  \centering
  \includegraphics[width=85mm]{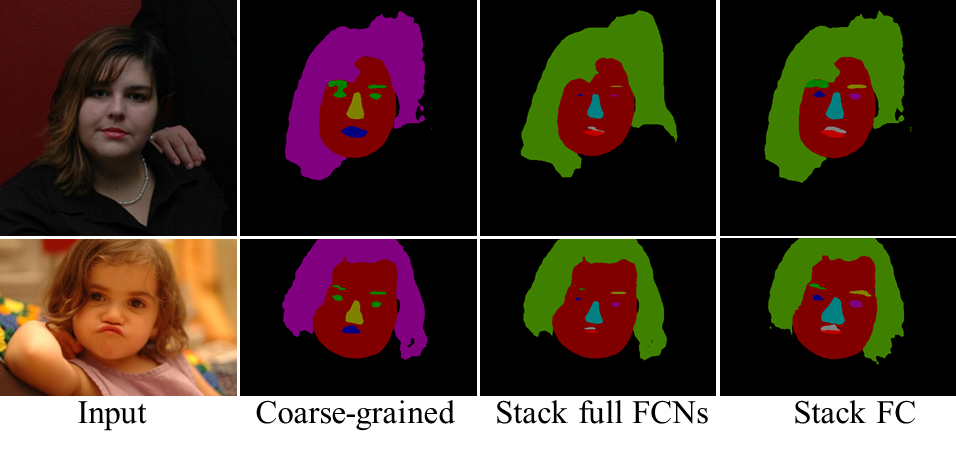}
  \caption{Comparison between stacking full FCNs and stacking FC modules. It is clear that the fine-grained result is more conform to the coarse-grained result when stacking FC.}
  \label{fig:cmp_full_fc}
\end{figure}

As shown in Fig.~\ref{fig:cmp_full_fc}, given the input image in the first column and the coarse-grained result in the second column, stacking a full FCN after it will feed the coarse-grained result to a deep network which results in inconsistency as shown in the third column. But by stacking only FC modules, the fine-grained result is more accord with the coarse-grained one as shown in the fourth column, so the following FC modules can pay more attention to parsing other details. It should be pointed out that in Fig.~\ref{fig:framework}, it shows a network stacking 3 FCs at 3 grain levels, but in fact our framework can be extended to more FC modules (2 to more).

\subsection{Skip connections for fine-grained parsing}

In a deep network, the receptive fields of deeper layers are larger, but the small structures and localization information at deep layers are lost due to the layer by layer convolution and pooling operations. To get better parsing result with small structures at the finer-grained level, we propose to add skip connections from the shallow layers in the network to the fine-grained FC modules. The finer the FC module is going to parsing, the shallower layers should be connected to.

\begin{figure}
  \centering
  \includegraphics[width=85mm]{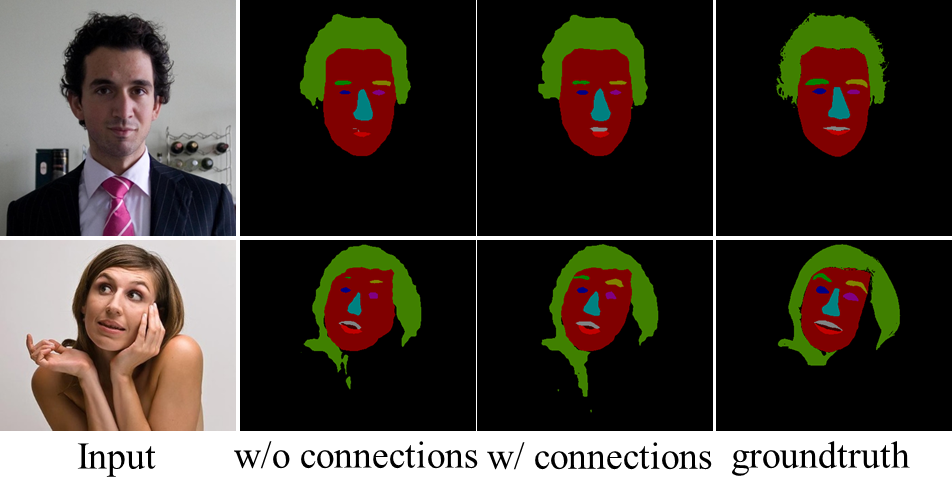}
  \caption{Comparison between with and without skip connections from shallow layers at the fine-grained parsing.}
  \label{fig:cmp_skip}
\end{figure}

To be specific, as shown in Fig.~\ref{fig:framework}, the first FC module parses images in a coarsest-grained level, so it only need to be connected to the network directly which composes a full FCN. For the medium-grained FC, it need to recognize some medium size structures. So we add a skip connection from medium layers to it. Similarly, the fine-grained FC should add connection from the shallow layers. That is to say, for any of the finer-grained FC modules (any FC except for the first one), the input comes from three: the feature map from the last layer of the network (before the FC modules), the score map from the former FC module, and the feature map from a shallow layer of the network. Given these three input, we first upsample the smaller feature maps to match the size of the largest feature map via bilinear interpolation, and then concatenate these features maps before feeding to that FC module. It can be formulated as follows:
\begin{equation}
{P_t} = {{\mathop{\rm F}\nolimits} _t}\left( {{\mathop{\rm Up}\nolimits} \left( {{f_0}} \right) \oplus {\mathop{\rm Up}\nolimits} \left( {{P_{t - 1}}} \right) \oplus {\mathop{\rm Up}\nolimits} \left( {{f_{ - t}}} \right)} \right) ,
\end{equation}
where ${{\mathop{\rm F}\nolimits} _t}$ represents a composite function of operations of the $t$-th FC module and $P_t$ is the output, $f_0$ is the feature map from the last layer of the network, $f_{-t}$ is the feature map from corresponding shallow layer, ${\mathop{\rm Up}\nolimits} \left(  \cdot  \right)$ upsample the smaller feature maps to match the size of the largest one, and $\oplus$ represent the concatenate operation. Fig.~\ref{fig:cmp_skip} illustrates the different performances between with and without skip connections from shallow layers. From it we can see that, after adding skip connections from shallow layers, the small structures (e.g., eyes, browns and mouth) can be parsed more precise.

\subsection{Training with coarse-to-fine supervision}

To train the stacked network, we define the loss function as a weighted combination of every individual grain level parsing after each FC module:
\begin{equation}
{{\cal L}_{total}} = \mathop \sum \limits_i {\lambda _i}{{\cal L}_i} ,
\end{equation}
where ${{\cal L}_i}$ is the loss at the $i$-th grained level image parsing, and ${\lambda _i}$ is the associated weight. We use equal weights, and apply the pixel-wise cross-entropy loss at every grain level in our experiments. Except for the last loss, others can be regarded as auxiliary losses which give coarse-to-fine hierarchical supervision help optimize the learning process.

For the training data, we use the fine-grained groundtruth label map to build the coarse-to-fine label maps. Specifically, we use the groundtruth label map directly for the last FC module (i.e., the finest-grained one), and then we merge some similar classes in the groundtruth label maps as coarse classes to get the coarse-grained training data for the penultimate FC. If there are more FC modules ahead, we continue to merge classes in the coarse-grained classes to further get coarser-grained training data. An example is shown in Fig.~\ref{fig:framework}, given the fine-grained groundtruth label map at the last FC module, the two coarser-grained label maps at the former two FC modules are built by merging classes in the groundtruth. A fundamental principle of this merging process is do not put small structures at coarse-grained label maps. Table~\ref{tab:parse_hierarchy} gives an example of our label merging rules on three dataset.

\newcommand{\tabincell}[2]{\begin{tabular}{@{}#1@{}}#2\end{tabular}}
\begin{table*}[t]
\begin{center}
\caption{The predefined three-grained parsing hierarchy on three public datasets.}
\label{tab:parse_hierarchy}
\small
\begin{tabular}{|l|l|l|l|}
  \hline
  ~ & \makebox[4.5cm][c]{HELEN Face~\cite{Smith2013c}} & \makebox[3.5cm][c]{PASCAL-Person-Parts~\cite{Chen2015c}} & \makebox[6cm][c]{ATR~\cite{Liang2015a}}  \\
  \hline
  \tabincell{c}{Coarse\\ grained} & \tabincell{l}{background, face, hair} & \tabincell{l}{background, upper body,\\ lower body} & \tabincell{l}{background, head, torso, lower body, bag+carf} \\
  \hline
  \tabincell{c}{Medium\\ grained} & \tabincell{l}{background, face skin, eyes, nose, mouth,\\ hair} & \tabincell{l}{background, head+torso,\\ arms, upper legs, lower legs} & \tabincell{l}{background, hat+hair, sunglass+face, upper-clothes,\\ skirt+pants+dress+belt, arms, shoes, leg, bag+carf } \\
  \hline
  \tabincell{c}{Fine\\ grained} & \tabincell{l}{background, face skin, left eyebrow,\\ right eyebrow, left eye, right eye, nose,\\ upper lip, inner mouth, lower lip, hair} & \tabincell{l}{background, head, torso,\\ upper arms, lower arms,\\ upper legs, lower legs} & \tabincell{l}{background, hat, hair, sun glass, upper-clothes,\\ skirt, pants, dress, belt, left-shoe, right-shoe, face,\\ left-leg, right-leg, left-arm, right-arm, bag, scarf} \\
  \hline
\end{tabular}
\end{center}
\end{table*}

\section{Experiments}

\subsection{Experimental settings}

\noindent\textbf{Dataset}. Since our algorithm is built to parsing images from coarse to fine, the test image must support hierarchy parsing. In this regard, we select a face parsing dataset and two person parsing datasets. For face parsing, we use the HELEN dataset~\cite{Smith2013c}. It is a face parsing dataset containing 2330 face images of arbitrary size, and each pixel is labeled as one of 11 classes. We use 2,000 images for training and the left for validation. For this dataset, we define the three grained levels of semantic classes as shown in Table~\ref{tab:parse_hierarchy}.

For person parsing, we use two datasets. The first is the merged PASCAL-Person-Parts~\cite{Chen2015c}, which contains six person part classes and one background class. It contains 3,533 images. We use the same setting of data splits as in~\cite{Chen2015c} with 1,716 images for training and the left for validation. Since it contains only 6 classes which is not so enough for a coarse-to-fine parsing, we add another dataset. That is the ATR dataset~\cite{Liang2015a} which contains as many as 18 person part classes. Totally, 7,700 images are included in the ATR dataset, with 6,000 for training, 1,000 for testing and 700 for validation. For these two datasets, we also build a three-grained parsing hierarchy as shown in Table~\ref{tab:parse_hierarchy}.

~\newline
\noindent\textbf{Network details}. In our experiments, we use the Deeplab-ResNet~\cite{Chen2017} without any tricks (\emph{incl.} multi-scale inputs, data augmentation, CRF and so on) as our base FCN. Layers after \emph{res5c} are regarded as FC module. For the three grain levels, the coarsest-grained FC module is directly connected to layer \emph{res5c}, and the other two are stacked one by one after the first FC. Besides, we add skip connections from layer \emph{res3b3} to the medium-grained FC module, and from layer \emph{res2c} to the fine-grained FC module.
To concatenate with these feature maps from shallow layers, we upsample the current feature map using bilinear interpolation to match the sizes.
Before training, we merge labels follows Table~\ref{tab:parse_hierarchy} to get coarse-to-fine label maps as the hierarchical supervision. That is to say, one training image has three corresponding label maps. We implement the proposed stacked networks based on the tensorflow implementation
 of Deeplab-ResNet, and fine-tune the pre-trained deeplab network for our coarse-to-fine image parsing.

\subsection{The effectiveness of proposed method}

We evaluate the effectiveness of proposed strategies on all three dataset using the pre-defined parsing hierarchy. For comparison, we also try to stack three full FCNs one after another to parse images from coarse to fine. To investigate the improvement over raw networks, for each grained level, we also train a standalone basic network (i.e., Deeplab-ResNet in this experiment) to parse images at that grained level. The performance of each strategy is measured in terms of mean pixel intersection-over-union over all classes (mIoU).

Table~\ref{tab:eff_helen}, Table~\ref{tab:eff_part} and Table~\ref{tab:eff_atr} show the results on each dataset respectively. It can be seen that stacking full FCNs nearly show no improvement in comparison with an individual FCN. But when selectively stacking the FC modules with coarse-to-fine hierarchical supervision, the improvement is significant for all three grained level. For HELEN and ATR dataset, adding skip connections from shallow layers to the finer-grained FC modules can further improve the performance of the network. This demonstrates the effectiveness of information from shallow layers for fine-grained parts parsing. But for PASCAL-Person-Parts dataset, adding skip connections has limited effects on the performance. We think this is because the PASCAL-Person-Parts dataset is a pre-merged dataset which have little inherent small structures, and the scale and location of parts in these images various greatly. This makes localization information from shallow layers can give limited help for parsing images in the PASCAL-Person-Parts dataset.

\begin{table}[t]
\begin{center}
\caption{Quantitative comparison of different strategies on HELEN dataset.}
\label{tab:eff_helen}
\small
\begin{tabular}{c|c|c|c}
  \hline
  mIoU & \tabincell{c}{Coarse\\ grained} & \tabincell{c}{Medium\\ grained} & \tabincell{c}{Fine\\ grained}  \\
  \hline
  Standalone & 0.766 & 0.645 & 0.478 \\
  \hline
  Stack full FCNs & 0.775 & 0.656 & 0.465 \\
  \hline
  Stack FC modules & 0.783 & 0.682 & 0.545 \\
  \hline
  \tabincell{c}{Stack FC modules with\\ skip connections} & 0.828 & 0.719 & 0.637 \\
  \hline
\end{tabular}
\end{center}
\end{table}

\begin{table}[t]
\begin{center}
\caption{Quantitative comparison of different strategies on PASCAL-Person-Parts.}
\label{tab:eff_part}
\small
\begin{tabular}{c|c|c|c}
  \hline
  mIoU & \tabincell{c}{Coarse\\ grained} & \tabincell{c}{Medium\\ grained} & \tabincell{c}{Fine\\ grained}  \\
  \hline
  Standalone & 0.743 & 0.593 & 0.564 \\
  \hline
  Stack full FCNs & 0.741 & 0.585 & 0.567 \\
  \hline
  Stack FC modules & 0.760 & 0.618 & 0.596 \\
  \hline
  \tabincell{c}{Stack FC modules with\\ skip connections} & 0.760 & 0.615 & 0.604 \\
  \hline
\end{tabular}
\end{center}
\end{table}

\begin{table}[t]
\begin{center}
\caption{Quantitative comparison of different strategies on ATR dataset.}
\label{tab:eff_atr}
\small
\begin{tabular}{c|c|c|c}
  \hline
  mIoU & \tabincell{c}{Coarse\\ grained} & \tabincell{c}{Medium\\ grained} & \tabincell{c}{Fine\\ grained}  \\
  \hline
  Standalone & 0.791 & 0.751 & 0.600 \\
  \hline
  Stack full FCNs & 0.799 & 0.739 & 0.575 \\
  \hline
  Stack FC modules & 0.804 & 0.766 & 0.632 \\
  \hline
  \tabincell{c}{Stack FC modules with\\ skip connections} & 0.809 & 0.772 & 0.653 \\
  \hline
\end{tabular}
\end{center}
\end{table}

\subsection{Compared with state-of-the-art}

In this section, we compare the proposed method with strong baselines on two human parsing datasets. We merge the groundtruth classes as shown in Table~\ref{tab:parse_hierarchy}, and train our stacked network with the coarse-to-fine label maps. The trained network is used to parsing images at the finest-grained level which is consistent with the groundtruth. The performance of our stacked network and some strong baseline models are compared.

For PASCAL-Person-Parts, we use the same setting of data splits as in~\cite{Chen2015c} with 1,716 images to train the network, and test the network over the left images. The performance is measured in terms of mIoU as in~\cite{Chen2015c}. Table~\ref{tab:cmp_part} shows the performance of our stacking network and comparisons with several state-of-the-art methods. Also, we have test the performance of our baseline network. From the table we can find that our baseline model can be obviously improved using the proposed stacking framework. And the stacked network can outperform the three state-of-the-art methods particularly.

For the ATR dataset, we train the network using 6,000 training images and evaluate the network over the 1,000 testing images. We use the same evaluation criterion as in~\cite{Liang2015a} and~\cite{Liang2015f}, including accuracy, accuracy over foreground classes, average precision, average recall, and average F-1 score over pixels. Table~\ref{tab:cmp_atr} shows the results. The stacking framework can improve the baseline network with significant gains. Except for the overall accuracy including background, the stacked network outperform all the three state-of-the-art methods over all other criterions by a wide margin. All these demonstrates that our stacking strategy is significantly helpful for human parsing tasks.

\begin{table}[t]
\begin{center}
\caption{Comparison with three state-of-the-art methods when evaluating on PASCAL-Person-Parts.}
\label{tab:cmp_part}
\small
\begin{tabular}{p{0.3\textwidth}|c}
  \hline
  \textbf{Method} & \textbf{mIoU(\%)}  \\
  \hline \hline
  Attention~\cite{Chen2015c} & 56.39 \\
  \hline
  LG-LSTM~\cite{Liang2015} & 57.97 \\
  \hline
  Attention+SSL~\cite{Gong2017} & 59.36 \\
  \hline \hline
  Deeplab (baseline) & 56.43 \\
  \hline
  Stacked Deeplab & \textbf{60.35} \\
  \hline
\end{tabular}
\end{center}
\end{table}

\begin{table}[t]
\begin{center}
\caption{Comparison with three state-of-the-art methods when evaluating on ATR dataset.} \label{tab:cmp_atr}
\small
\begin{tabular}{l|c|c|c|c|c}
  \hline
  ~ & acc. & \tabincell{c}{f.g.\\acc.} & \tabincell{c}{avg.\\prec.} & \tabincell{c}{avg.\\recall} & \tabincell{c}{avg.\\F-1} \\
  \hline \hline
  M-CNN~\cite{Liu2015d} & 89.57 & 73.98 & 64.56 & 65.17 & 62.81 \\
  \hline
  ATR~\cite{Liang2015a} & 91.11 & 71.04 & 71.69 & 60.25 & 64.38 \\
  \hline
  Co-CNN~\cite{Liang2015f} & \textbf{95.23} & 80.90 & 81.55 & 74.42 & 76.95 \\
  \hline \hline
  Deeplab (baseline) & 93.88 & 79.80 & 80.71 & 69.47 & 73.38 \\
  \hline
  Stacked Deeplab & 95.01 & \textbf{83.50} & \textbf{82.66} & \textbf{76.61} & \textbf{79.08} \\
  \hline
\end{tabular}
\end{center}
\end{table}

\subsection{The generalization of proposed stacking strategies}

In this experiment, we will evaluate the generalization of the proposed stacking strategies for other networks. We try to inject our coarse-to-fine stacked framework into several popular image segmentation networks, including SegNet~\cite{Badrinarayanan2015}, FC-DenseNet~\cite{Jegou2017} and PSPNet~\cite{Zhao2016a}, and compare the performance of the raw network and the stacked network on HELEN dataset. We define the coarse-to-fine parsing hierarchy as shown in Table~\ref{tab:parse_hierarchy}. For the raw networks, we train them with the finest-grained label maps. And for the stacked networks, we train them with the coarse-to-fine label maps. The finest-grained parsing performances are measured in terms of mIoU and compared.

For SegNet, we use the VGG16-based SegNet, and regard layers after layer \emph{upsample1} (i.e., the last upsampling layer) excluding the \emph{Softmax} layer as the FC module. We directly add two additional FC modules after the raw FC in the network, and use these three FC for the coarse-to-fine parsing. Since the raw SegNet has several pooling indices which connect the shallow layers to the deeper layers firmly, we donot add skip connections anymore. For FC-DenseNet, we use the FC-DenseNet56 as our base network, and regard the last convolution layers as the FC module. We add skip connections from the third dense block to the medium-grained FC module and from the first dense block to the fine-grained FC module. For PSPNet, we use the ResNet18-based PSPNet for simplicity, and regard the convolution layers after the pyramid pooling module as the FC module. Two additional FC modules are added after the raw FC in the network. Skip connections from the third and the first residue blocks are added to the second and third FC modules respectively.

\begin{table}[t]
\begin{center}
\caption{Comparison between some popular FCN and their stacked version using proposed method. The performance is measured on HELEN dataset in terms of mIoU.} \label{tab:generalization}
\small
\begin{tabular}{p{0.3\textwidth}|c}
  \hline
  \textbf{Method} & \textbf{mIoU(\%)}  \\
  \hline\hline
  SegNet~\cite{Badrinarayanan2015} & 57.85 \\
  \hline
  Stacked SegNet & 62.94 \\
  \hline\hline
  FC-DenseNet~\cite{Jegou2017} & 49.80 \\
  \hline
  Stacked FC-DenseNet & 53.28 \\
  \hline\hline
  PSPNet~\cite{Zhao2016a} & 49.40 \\
  \hline
  Stacked PSPNet & 52.65 \\
  \hline
\end{tabular}
\end{center}
\end{table}

Table~\ref{tab:generalization} shows the result. The proposed stacking strategies can improve existing image segmentation networks significantly: 5.09\% for SegNet, 3.48\% for FC-DenseNet and 3.25\% for PSPNet. The improvements for SegNet and FC-DenseNet indicate that with the context prior from the coarse-grained result, image parsing can be done better. And for PSPNet, although the pyramid pooling module can provide effective global contextual prior, our coarse-to-fine stacking framework can provide more contextual information to further improve it.

\section{Conclusion}
In this paper, we propose a coarse-to-fine image parsing framework to parse images progressively with refined semantic classes by stacking the segmentation modules in off-the-shelf semantic segmentation networks. The former segmentation module is used to parse images at a coarser-grained level, and the result will be feed to the following one to provide effective contextual clues for the finer-grained parsing. For finer-grained parsing modules, we add skip connections from shallow layers in the network to recover details of small structures. We merge classes in groundtruth to get coarse-to-fine label maps, and train the stacked network with these hierarchical supervision end-to-end. This type of coarse-to-fine parsing framework can be injected into many advanced neural networks to improve the results. Extensive evaluations demonstrate the effectiveness of our method.

\ninept

\end{document}